\documentclass[letterpaper, 10 pt, conference]{ieeeconf}  

\IEEEoverridecommandlockouts                              

\usepackage{epsfig} 
\usepackage{amsmath} 
\usepackage{amssymb}  

\usepackage{bm}
\usepackage{caption}
\usepackage{subcaption}
\usepackage{xcolor}
\usepackage{physics}
\usepackage{hyperref}

\renewcommand{\baselinestretch}{0.96}

\title{{\scriptsize This work has been submitted to the IEEE for possible publication. Copyright may be transferred without notice, after which this version may no longer be accessible.}
	
	\LARGE \bf
	Robot Control for Simultaneous Impact Tasks \\ through Time-Invariant Reference Spreading
}

\author{Jari J. van Steen, Nathan van de Wouw and Alessandro Saccon
}
\begin{document}

\maketitle
\thispagestyle{empty}
\pagestyle{empty}

	\begin{abstract}

	With the goal of enabling the exploitation of impacts in robotic manipulation, a new framework is presented for control of robotic manipulators that are tasked to execute nominally simultaneous impacts. 
	In this framework, we employ tracking of time-invariant reference vector fields corresponding to the ante- and post-impact motion, increasing its applicability over similar conventional tracking control approaches. The ante- and post-impact references are coupled through a rigid impact map, and are extended to overlap around the area where the impact is expected to take place, such that the reference corresponding to the actual contact state of the robot can always be followed. 
	As a sequence of impacts at the different contact points will typically occur, resulting in uncertainty of the contact mode and unreliable velocity measurements, a new interim control mode catered towards time-invariant references is formulated. In this mode, a position feedback signal is derived from the ante-impact velocity reference, which is used to enforce sustained contact in all contact points without using velocity feedback. 
	With an eye towards real implementation, the approach is formulated using a QP control framework, and is validated using numerical simulations both on a rigid robot with a hard inelastic contact model and on a realistic robot model with flexible joints and compliant partially elastic contact model.
	
	\end{abstract}

	\section{Introduction}\label{sec:introduction}

	Exploitation of intentional impacts plays a vital role in increasing the versatility and efficiency of robots for a range of applications in locomotion and manipulation. In locomotion, such intentional impacts enable humanoid running \cite{Tajima2009}, while for manipulation, utilizing intentional impacts can ensure a faster and more energy-efficient handling of objects \cite{Stouraitis2020}, which increases the applicability of robots in an industrial setting. While exploiting impacts is part of the repertoire of human skills, seen in applications like running or kicking a ball, such impacts can cause issues when translated to applications in robotics. Clearly, hardware can break when impact forces are too high. But even when impacts stay within safe limits such that no damage occurs, the rapid velocity transitions that result from the impact can result in performance and stability issues when controlling the robot if not accounted for. An example of the latter is the peak in the error between the desired and actual velocity of the robot under tracking control, which appears as a result of the inevitable mismatch in time between the expected velocity jump in the reference, and the actual velocity jump \cite{Biemond2013,Leine2008}. 
	
	This work will focus specifically on what we call nominally simultaneous impacts, which take place when multiple points between the robot and its environment are tasked to simultaneously come into contact at non-zero speed. An example of an application that shows such behavior is dual arm industrial robots swiftly grabbing boxes \cite{Benali2018}. As the impacting surfaces are likely not perfectly aligned at the time of impact, a series of unplanned intermediate impacts and a corresponding unpredictable series of velocity jumps will typically occur instead of the planned nominal simultaneous impact. As we will show, this implies that velocity feedback control cannot be reliably used during an impact sequence.
	
	In recent years, a handful of control methods have been developed to accurately execute motions that include simultaneous impacts while avoiding unwanted spikes in the control inputs. 
	This includes tracking approaches such as \cite{Yang2021,Morarescu2010}, and the framework of reference spreading, introduced in \cite{Saccon2014} and expanded upon in \cite{Rijnen2015,Rijnen2019a,Rijnen2020}.  
	Reference spreading is a hybrid control approach that deals with impacts by defining ante- and post-impact references that are coupled by an impact map at the nominal impact time, and overlap about this time. It is ensured that the tracking reference corresponds with the actual contact state of the robot by switching the reference based on impact detection, which avoids error peaking and related spikes in the control inputs. 
	While initially defined for single impacts, reference spreading has been shown to lend itself for tracking control for scenarios with inelastic simultaneous impacts in \cite{Rijnen2019, Steen2022}. This is achieved by including an interim control mode in which velocity error feedback is turned off, which is active from the moment of the first impact until full contact is established. 
	
	All of the aforementioned approaches rely on tracking control of trajectories that are explicitly time-dependent. While this can be used to accurately execute desired motions and make impacts with a desired velocity at a desired time, it is not necessarily an ideal strategy for all applications \cite{Hauser1995,Aguiar2005}. 
	If collision avoidance, conflicting tasks or an unexpected disturbance cause the robot to lag behind the nominal motion, it is generally undesirable to force the robot back to a trajectory prescribed in time. Additionally, it is desirable to prescribe a task in a way that is robust to variations in the initial conditions, such that it is not required to replan a trajectory when the initial robot pose is altered. 
	
	A time-invariant control approach, where the robot follows a certain desired velocity reference based on its current posture, such as \cite{Khansari-Zadeh2012b}, would be more suitable for such scenarios. 
	However, time-invariant control schemes that deal with motions containing (simultaneous) impacts are limited. State-of-the-art time-invariant robot control approaches such as \cite{Salehian2018} focus on impact avoidance rather than exploitation of impacts. Previous works have focused on using learned time-invariant references to hit objects with nonzero velocity \cite{Khansari-Zadeh2012,Khurana2021}. However, potential unwanted behavior resulting directly from the impact is not explicitly addressed here.
	
	The contribution in this paper is the formulation of a time-invariant approach, used for control of mechanical systems tasked to perform motions containing nominally simultaneous inelastic impacts. The proposed controller has a switching structure that resembles the controller proposed in \cite{Steen2022}, which consists of an ante-impact, interim, and post-impact control mode.	We also show that the new time-invariant approach can be cast into the quadratic programming (QP) robot control framework \cite{Bouyarmane2019,Salini2010, Escande2012}, allowing to include constraints ensuring for example collision avoidance and adherence to joint limits, which are essential in real applications. Impact-aware QP control is an active area of research \cite{Dehio2021, Wang2019}, and this paper contributes to that line. 
	While \cite{Steen2022} uses the classical time-based formulation of reference spreading, where the ante- and post-impact references overlap in time about the nominal impact time, this paper introduces the notion of \textit{time-invariant reference spreading}, in which configuration-dependent ante- and post-impact reference velocity and force fields are formulated. These fields overlap in position about the set of locations where the impact is expected to occur, and are coupled through a nominal impact map at these locations. 
	As \cite{Steen2022} relies on position feedback in the interim mode, a new approach based on time integration of the velocity references is proposed due to the lack of an explicit position reference in the time-invariant reference spreading approach.

	The structure of this paper is as follows. In Section \ref{sec:robot_dynamics}, we will provide the equations of motion of the 3DOF robot that we will use to demonstrate the proposed control approach. 
	In Section \ref{sec:reference_generation}, the generation of the ante- and post-impact reference velocity fields
	is described. Section \ref{sec:control_approach} presents the control framework consisting of the ante-, interim-, and post-impact modes, used to track these references. 
	In Section \ref{sec:numerical_validation}, we validate the approach against two other baseline methods by means of numerical simulations, before drawing conclusions in Section \ref{sec:conclusion}.

	\section{Robot dynamics}\label{sec:robot_dynamics}

\begin{figure}
		\centering
		\includegraphics[width=\linewidth]{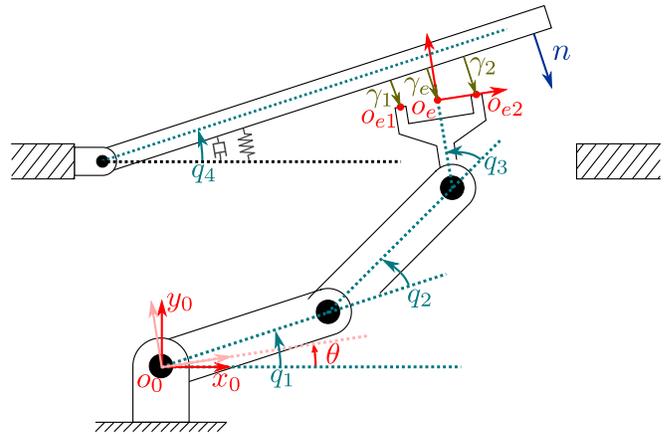}
		\caption{Overview of the 3DOF planar manipulator impacting a hinged rigid plank.}\label{fig:robot_3DOF}
	\end{figure}

	While the control strategy described in this paper can be applied to a wider range of scenarios, the planar manipulator depicted in Figure \ref{fig:robot_3DOF}, which is identical to the manipulator used throughout \cite{Steen2022}, is used in this paper to illustrate and demonstrate the approach.  
	The robot consists of three rigid frictionless actuated joints and rigid links, and impacts a hinged rigid plank. The system's generalized coordinate vector is $\bm q = \left[\bm q_\text{rob}^T \ q_4\right]^T$ with $\bm q_\text{rob} = \left[q_1 \ q_2 \ q_3\right]^T$. The mass of each link $i$ and its inertia around the center of gravity are given by $m_i$ and $I_{g,i}$ respectively, with the inertia of the plank around the hinge given by $I_{o,4}$. The vector normal to the plank is described by $\bm n$. 
	The end effector position $\bm p$ and orientation $\theta$ describe the position and orientation of frame $o_e x_ey_e$ expressed in terms of frame $o_0x_0y_0$, with their respective velocities given by 
	\begin{equation}
	\dot{\bm p} = \bm J_p(\bm q) \dot{\bm{q}}, \ \ \ \dot{\theta} = \bm J_\theta(\bm q) \dot{\bm{q}}.
	\end{equation}
	with $\bm J_p(\bm q) = \left[\bm J_{p,\text{rob}}(\bm q) \ \bm 0\right]$, $\bm J_\theta(\bm q) = \left[\bm J_{\theta,\text{rob}}(\bm q) \  0\right]$, and
	$$
	\bm J_{p,\text{rob}}(\bm q) = \frac{\partial \bm p}{\partial \bm q_\text{rob}},	\ \ \	\bm J_{\theta,\text{rob}}(\bm q) = \frac{\partial \theta}{\partial \bm q_\text{rob}},
	$$
	For ease of notation, in the following we will drop the explicit dependency on $\bm q$ (or $\dot{\bm q}$). The contact between the end effector and the plank can occur at two distinct point as shown in Figure \ref{fig:robot_3DOF}, and is assumed to be frictionless. These contacts can be described by the gap functions $\gamma_i$ and corresponding contact forces $\lambda_i$, for $i \in\{1,2\}$, satisfying the complementarity conditions 
	\begin{equation}\label{eq:complementarity}
	0 \leq \gamma_i \perp \lambda_i \geq 0,
	\end{equation}
	as is common in the framework of nonsmooth mechanics \cite{Brogliato2016}. This implies that the contact force is zero when the contact is open and non-negative when the contact is closed.
	\subsection{Non-impulsive free and constrained dynamics}
	The robot dynamics in free and constrained motion are
	\begin{equation}\label{eq:EOM}
	\bm M \ddot{\bm q} + \bm h = \bm S \bm \tau + \bm J_{N}^T  \bm \lambda,
	\end{equation}
	with joint positions $\bm q$, mass matrix $\bm M$, vector of gravity, centrifugal, and Coriolis terms $\bm h$, applied joint torques $\bm \tau$, actuation matrix $\bm S = \left[\bm I_{3 \times 3} \ \bm 0_3 \right]^T$, normal contact forces $\bm \lambda$, and contact Jacobian $\bm J_N = \left[\bm J_{N,1}^T \ \bm J_{N,2}^T\right]^T$, with
	\begin{equation}
	\bm J_{N,i} = \frac{\partial \gamma_i}{\partial \bm q}.
	\end{equation}
	
	\subsection{Impulsive impact dynamics}

	To ensure adherence to the non-penetration condition $\gamma_i \geq 0$ in \eqref{eq:complementarity}, an instantaneous jump in $\dot{\bm q}$ has to be allowed as soon as a contact is closed. This state-dependent jump is described by a so-called impact map \cite{Glocker2006}, causing a discontinuity in joint velocities at the impact time, with ante-impact velocity $\dot{\bm q}^-$, post-impact velocity $\dot{\bm q}^+$, and continuous position $\bm q^- = \bm q^+$. Integrating \eqref{eq:EOM} over the impact time with $\lambda$ allowed to become impulsive, leads to the impact equation \cite{Brogliato2016} as
	\begin{equation}\label{eq:impact_eq_new}
	\bm M\left(\dot{\bm q}^+ - \dot{\bm q}^{-}\right) = \bm J_{N}^T \bm \Lambda,
	\end{equation}
	with $\bf{\bm \Lambda} \in \mathbb{R}^2$ the momentum associated to the impulsive contact forces. We assume  inelastic impacts to take place at the moment of contact transition, which means we assume that the impact law is given by $\dot{\gamma}_i^+ = 0$. This means that, when contact of the end effector with the plank is established simultaneously at both possible contact locations, we have 
	\begin{equation}\label{eq:impact_law_new}
	\bm J_{N} \dot{\bm q}^+ = \bm 0.
	\end{equation}	
	Pre-multiplying \eqref{eq:impact_eq_new} by $\bm J_N \bm M^{-1}$ and substituting \eqref{eq:impact_law_new} gives
	\begin{equation}\label{eq:impact_map_derivation}
	-\bm J_N^T \dot{\bm q}^- = \bm J_N \bm M^{-1} \bm J_{N}^T \bm \Lambda.
	\end{equation}
	From \eqref{eq:impact_map_derivation}, $\bm \Lambda$ is extracted and substituted in \eqref{eq:impact_law_new}, leading to the simultaneous impact map 
	\begin{equation}\label{eq:impact_map} 
	\dot{\bm q}^+ = \left(\bm I - \bm M^{-1} \bm J_{N}^T \bm \left(\bm J_{N} \bm M^{-1} \bm J_{N}^T\right)^{-1}\bm J_{N}\right)\dot{\bm q}^{-},
	\end{equation}
	which will be used in the formulation of the reference velocity fields in the next section.
	
	\section{Desired path generation} \label{sec:reference_generation}
	
	In this section, we will give details on the design of the time-invariant ante-impact and post-impact reference velocity fields as a function of the end-effector pose, which we use in the control law in Section \ref{sec:control_approach} to execute the desired motion.

	\subsection{Ante-impact}\label{sec:reference_generation_ante}
	
	The process of generating the desired ante-impact velocity can be decoupled into two parts, the first being a reference to guide the end effector position towards the hinged plank with a given velocity denoted by $\bar{\dot{\bm p}}_d^a(\bm p)$, with $\bar{\cdot}$ indicating the extension of the reference. The second part is the orientation reference $\dot{\theta}_d^a(\theta)$ to align the orientation of the end effector with the hinged plank to pursue a simultaneous impact. 
	
	\subsubsection{Position}
	
	The ante-impact linear velocity reference $\bar{\dot{\bm p}}_d^a(\bm p)$ is formulated with the goal of guiding the end effector towards the line or area where the desired impact is expected to take place, which is defined as the \emph{nominal impact line}. In our use case, this nominal impact line is given by the bottom of the plank with the plank in a fixed a priori estimated equilibrium state $(q^-_{4,\text{est}}, \dot{q}^-_{4,\text{est}}) = (0,0)$ at the time of impact.  
	
\begin{figure}
		\centering
		\includegraphics[width=0.94\linewidth]{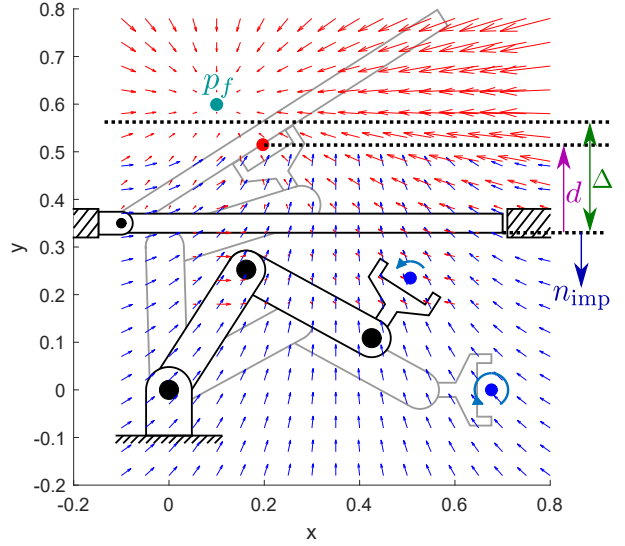}
		\caption{Depiction of the ante- and post-impact linear velocity references $\bar{\dot{\bm p}}^a_{d}$ (blue) and $\bar{\dot{\bm p}}^p_{d}$ (red), and ante-impact angular velocity reference $\dot{{\theta}}^a_{d}$ (blue) with an initially stationary plank.} 
		\label{fig:reference}
	\end{figure}

	We define $d: \mathbb{R}^2 \to  \mathbb{R}, p \mapsto d(\bm p)$ as the smallest distance from the point $\bm p$ to this nominal impact line, as shown in Figure \ref{fig:reference}. With this, the first step to formulate $\bar{\dot{\bm p}}_d^a(\bm p)$ is the formulation of a velocity reference $\dot{\bm p}_d^a(\bm p)$ for all end effector positions $\bm p$ that lie before the nominal impact line, corresponding to $d(\bm p) < 0$, which is formulated in Appendix A. 
	In line with the core idea of reference spreading, we extend this reference past the nominal impact line to formulate the extended velocity reference given by 
	\begin{equation}\label{eq:extension_ante}
	\bar{\dot{\bm p}}_d^a(\bm p) = \left\{
	\begin{aligned}
	{\dot{\bm p}}_d^a(\bm p) \quad & \text{if} \ d(\bm p) \leq 0,  \\
	{\dot{\bm p}}_d^a(\bm p + d(\bm p) \bm n_\text{imp}) \quad & \text{if} \ d(\bm p) > 0,
	\end{aligned}
	\right.
	\end{equation}
	with $\bm n_\text{imp}$ as the normal vector $\bm n$ corresponding to the estimated plank state $\bm q^-_{4,\text{est}}$ at the moment of impact as seen in Figure \ref{fig:reference}. This implies that for all positions $\bm p$ behind the nominal impact line, $\bar{\dot{\bm p}}^a_d$ is equal to the value of $\dot{\bm p}^a_d$ on the nominal impact line closest to $\bm p$. 
	This ensures that the controller always has an ante-impact reference to follow, even when the impact occurs at a different location than expected due to uncertainties in the environment.

	\subsubsection{Orientation}
	
	To enforce a simultaneous impact with the manipulator, an orientation task is formulated such that the orientation of the end effector $\theta$ aligns with the estimated ante-impact orientation of the plank, indicated by $q^-_{4,\text{est}}$. Hence, the desired orientation field is given by 
	\begin{equation}
	\dot{\theta}^a_{d}(\theta) =  K_\theta (q^-_{4,\text{est}} - \theta),
	\end{equation}
	with user-defined parameter $K_\theta \in \mathbb{R}^+$.
	
	\subsection{Post-impact}\label{sec:reference_generation_post}
	
	As opposed to the ante-impact reference, the post-impact reference only consists of a linear reference velocity field $\bar{\dot{\bm p}}_d^p(\bm p)$, as the orientation of the manipulator is implicitly described by this position, assuming the manipulator remains in contact with the plank at both contact points. This reference is designed with the goal of reaching the target final position $\bm p_f$, see Figure \ref{fig:reference}, 
	while also ensuring that the post-impact velocity is compatible with the ante-impact velocity, resulting in minimal tracking error after the impact, and minimal resulting spikes in the control inputs.
	
	In order to achieve both goals, $\bar{\dot{\bm p}}_d^p(\bm p)$ is obtained by blending two vector fields. The first vector field, denoted by $\dot{\bm p}_{d,f}^p(\bm p)$, ensures $\bm p_f$ is reached, and is given by
	\begin{equation}\label{eq:p_d_f_p}
	\dot{\bm p}_{d,f}^p(\bm p) = k_f \left( \bm p_f - \bm p \right),
	\end{equation}
	with user-defined constant $k_f \in \mathbb{R}^+$. 
	The second vector field, $\bar{\dot{\bm p}}_{d,\text{imp}}^p(\bm p)$, ensures that the ante- and post-impact references are compatible by coupling $\bar{\dot{\bm p}}_{d,\text{imp}}^p(\bm p)$ to the ante-impact velocity reference ($\bar{\dot{\bm p}}_{d}^a(\bm p)$, $\dot{\theta}^a_d(\bm p)$) along the nominal impact line. This is achieved by using the simultaneous impact map \eqref{eq:impact_map} to determine the nominal post-impact velocities along the nominal impact line, and extending these velocities along the directions normal to the nominal impact line.
	A detailed formulation of $\bar{\dot{\bm p}}_{d,\text{imp}}^p(\bm p)$ is given in Appendix B. 
	A convex combination of the two reference velocity fields is taken to construct the desired extended post-impact velocity, as
	\begin{equation}
	\bar{\dot{{\bm p}}}^{p}_{d}(\bm p) = \left\{
	\begin{aligned}
	\bar{\dot{\bm p}}^{p}_{d,\text{imp}}(\bm p) \ & \text{if} \ \ \ \hspace{0.15cm} \  \ \ d(\bm p) \leq 0,  \\
	\begin{aligned}
	\frac{d(\bm p)}{\Delta} \bar{\dot{\bm p}}^{p}_{d,\text{imp}}(\bm p) \ + &  \\ \left(1 - \frac{d(\bm p)}{\Delta}\right) \dot{{\bm p}}^{p}_{d,f}(&\bm p )  \end{aligned} \ & \text{if} \ \ 0 < d(\bm p) < \Delta, \\
	\dot{{\bm p}}^{p}_{d,f}(\bm p) \ & \text{if} \ \Delta \leq d(\bm p),
	\end{aligned}
	\right.
	\end{equation}
	with $\Delta \in \mathbb{R}^+$ as the user-defined width of the band where both $\bar{\dot{\bm p}}^{p}_{d,\text{imp}}$ and $\dot{{\bm p}}^{p}_{d,f}$ contribute to the value of $\bar{\dot{\bm p}}^{p}_{d}$, as shown in Figure \ref{fig:reference}. For the nominal motion, this implies that the tracking error will be zero upon switching to the post-impact reference at the moment of impact. 
	Provided that 
	\begin{equation}
	\bm n_\text{imp}^T \bar{\dot{{\bm p}}}^{p}_{d}(\bm p_i) < 0, \forall \bm p_i  \text{ s.t. } 0 \leq d(\bm p_i) \leq \Delta,
	\end{equation}
	and
	\begin{equation}
	d(\bm p_f) > \Delta,
	\end{equation}
	it can be ensured that following the desired velocity reference $\dot{{\bm p}}^{p}_{d}$ implies that the manipulator moves away from the nominal impact line until $d(\bm p) > \Delta$, after which $\dot{\bm p}^p_d = \dot{{\bm p}}^{p}_{d,f}$, leading to the target position $\bm p_f$ to be reached in equilibrium.

	\section{Control approach}\label{sec:control_approach}
	
	Using the time-invariant reference velocity fields obtained from Section \ref{sec:reference_generation}, we show how to construct a control action for the ante-impact, interim and post-impact mode. 
	In the ante-impact and post-impact mode, the goal is to track the corresponding desired reference velocity field, with an additional force task in the post-impact mode. 
	The interim mode is active as long as contact is only partially established. Its main goal is completing the contact without using any velocity error feedback as a result of uncertainties in the contact state. The switching policy is based on detection of the first impact, which activates the interim mode, and then monitoring when full contact is established, which activates the post-impact mode. 
	Each of the modes has a corresponding discrete-time QP controller which is used to obtain a desired joint torque $\bm \tau^*$ to be applied to the robot after every fixed time step $\Delta t$. 
	
	Compared to \cite{Steen2022}, where a reference is explicitly described in time, the time-invariant nature of the proposed approach adds robustness for different initial conditions or large disturbances as explained in Section \ref{sec:introduction}. However, the time-invariant nature of the proposed approach implies that no explicit ante-impact position reference exists, while in \cite{Steen2022}, a feedback term based on the ante-impact position reference plays a vital role in the interim mode to establish full contact in the absence of reliable velocity feedback. To overcome this challenge, a renewed interim mode is designed that uses the ante-impact velocity reference designed in Section \ref{sec:reference_generation_ante} to construct a position feedback signal, as will be further clarified in Section \ref{sec:int_mode}.

	\subsection{Ante-impact mode}\label{sec:ante}
	
	In the ante-impact mode, the QP optimization variables are given by the input torques $\bm \tau$ we seek to obtain, and the joint accelerations $\ddot{\bm q}$, which are included to simplify the formulation of the cost function. 
	Through position measurements and velocity estimations, $\bm q$ and $\dot{\bm q}$ at each time step are assumed to be known. 
	The cost function of the ante-impact QP is a weighted sum of the costs corresponding to a position and orientation task, defined such that the ante-impact linear and angular velocity references $\bar{\dot{\bm p}}^a_{d}(\bm p)$ and $\dot{\theta}^a_{d}(\theta)$ defined in Section \ref{sec:reference_generation_ante} are followed. The error corresponding to tracking the linear velocity is formulated as
	\begin{equation}\label{eq:e1}
	\bm e^a_p = \ddot{\bm p} - \bar{\ddot{{\bm p}}}^a_{d}(\bm p) + k^a_p \left(\bar{\dot{{\bm p}}}^a_{d}(\bm p) - \bm J_p \dot{\bm q}\right),
	\end{equation}
	with gain $k^a_p \in \mathbb{R}^+$ and desired acceleration $\bar{\ddot{\bm p}}^a_{d}(\bm p)$ derived from $\bar{\dot{{\bm p}}}^a_{d}(\bm p)$ through 
	\begin{equation}\label{eq:ddot_p_a_d}
	\bar{\ddot{\bm p}}^a_{d}(\bm p) = \frac{\partial \bar{\dot{{\bm p}}}^a_{d}}{\partial \bm p} \bar{\dot{{\bm p}}}^a_{d}(\bm p).
	\end{equation}
	Enforcing $\bm e^a_p = 0$ implies that the desired closed-loop behavior of following the reference $\dot{{\bm p}}^a_{d}(\bm p)$ is indeed imposed. 
	Rewriting the error as 
	\begin{equation}\label{eq:pos_task_ante}
	\bm e^a_p = \bm J_{p}\ddot{\bm q} + \bm\eta_p^a,
	\end{equation}
	with
	\begin{equation}\label{eq:eta_p}
	\bm\eta^a_p = \dot{\bm J}_p\dot{\bm q} - \bar{\ddot{{\bm p}}}^a_{d}(\bm p) - k^a_p \left(\bar{\dot{{\bm p}}}^a_{d}(\bm p) - \bm J_p\dot{\bm q}\right),
	\end{equation}
	and similarly defining an error corresponding to the orientation task with gain $k_\theta$, leads to the weighted cost function
	\begin{equation}\label{eq:cost_ante}
	\begin{aligned}
	E_\text{ante} = & w_p\left(\ddot{\bm q}^T \bm J_{p}^T \bm J_{p} \ddot{\bm q} + 2\left.\bm \eta^a_p\right.^T \bm J_{p} \ddot{\bm q}\right) \\ + & w_\theta\left( \ddot{\bm q}^T \bm J_{\theta}^T \bm J_\theta\ddot{\bm q} + 2\eta^a_\theta \bm J_\theta \ddot{\bm q}\right),
	\end{aligned}
	\end{equation}
	with user-defined task weights $w_p, w_\theta \in \mathbb{R}^+$. Note that the terms independent of the optimization variables have been discarded in \eqref{eq:cost_ante}. Combining this cost function with the equations of motion of the free-moving system, corresponding to \eqref{eq:EOM} with $\bm \lambda = 0$, and a constraint forcing the torque to remain between lower and higher bounds $\underline{\bm \tau}$ and $\bar{\bm \tau}$, respectively, gives the full ante-impact QP as
	\begin{equation}
	(\ddot{\bm q}^*, \bm \tau^*) = \underset{\ddot{\bm q}, \bm \tau}{\operatorname{argmin}} \ E_\text{ante},
	\end{equation}
	s.t.
	\begin{align}\label{eq:EOM_ante}
	\bm M\ddot{\bm q} + & \ \bm h = \bm S \bm \tau, \\
		\underline{\bm \tau} \leq & \ \bm \tau \leq \bm \bar{\bm \tau}.
	\end{align}
	The reference torque $\bm \tau^*$ is then sent to the robot at all times $t_k$ separated by time step $\Delta t$, with $\bm q = \bm q(t_k)$, $\dot{\bm q} = \dot{\bm q}(t_k)$.
			
	\subsection{Interim mode}\label{sec:int_mode}

	As advocated earlier, we cannot rely on velocity feedback control using either the ante-, or post-impact velocity reference during the interim mode, as on a real robot the exact contact state is not known. 
	The challenge of this interim mode is hence to formulate a QP tasked with establishing full contact without relying on velocity error feedback. 
	
	Our approach aims to achieve this by applying torque as if the system is still in the ante-impact mode, while adding position feedback based on a position error constructed online using the ante-impact velocity reference and the pose detected at the first impact time. For the position, this reference is determined by time integration through
	\begin{equation}
		\bm p_d^\text{int}(t) = \bm p(t_\text{int}) + \int_{t_\text{int}}^{t}\bar{\dot{\bm p}}^a_d(\bm p) \ \text{d}t,
	\end{equation}
	with $t_\text{int}$ as the moment at which the interim mode starts. This can be iteratively approximated after each time step $\Delta t$ to obtain
	\begin{equation}\label{eq:pos_int}
	\bm p_d^\text{int}(t_k + \Delta t) = \bm p_d^\text{int}(t_k) + \bar{\dot{\bm p}}^a_d(\bm p_d^\text{int}(t_k)) \Delta t.
	\end{equation}

	To avoid velocity error feedback in the interim mode, we can, assuming that the manipulator is never in a singular configuration, replace $\dot{\bm q}$ in \eqref{eq:pos_task_ante} by the nominal joint velocity ${\dot{\bm q}}_\text{int}$ corresponding to the ante-impact velocity references $\bar{\dot{\bm p}}^a_{d}(\bm p)$ and $\dot{{\theta}}^a_{d}(\theta)$ as given by
	\begin{equation}\label{eq:q_itmd}
	{\dot{\bm q}}_\text{int} = \left[\left.{\dot{\bm q}}_{\text{rob},\text{int}}\right.^T \dot{q}^-_{4,\text{est}}\right]^T,
	\end{equation}
	with
	\begin{equation}\label{eq:int_task_int}
	{\dot{\bm q}}_{\text{rob},\text{int}} := \begin{bmatrix}\bm J_{p,\text{rob}} \\ \bm J_{\theta,\text{rob}}\end{bmatrix}^{-1}  \begin{bmatrix} \bar{\dot{\bm p}}^a_{d}(\bm p) \\ \dot{{\theta}}^a_{d}(\theta)\end{bmatrix},
	\end{equation}
	and $\dot{q}^-_{4,\text{est}}$ given by the estimated ante-impact velocity of the plank. 
	Replacing $\dot{\bm q}$ in \eqref{eq:pos_task_ante} by $\dot{\bm q}_\text{int}$ and adding position feedback using \eqref{eq:pos_int}, the interim mode cost is formulated as
	\begin{equation}\label{eq:pos_task_int}
	{\bm e}^\text{int}_{p} = \bm J_{p}\ddot{\bm q} + {\bm\eta}_{p}^\text{int},
	\end{equation}
	with
	\begin{equation}\label{eq:eta_int}
	{\bm\eta}_{p}^\text{int} = {\dot{\bm J}}_{p,\text{int}}{\dot{\bm q}_\text{int}} - \bar{\ddot{\bm p}}^a_{d}(\bm p) - k^\text{int}_{p} \left({\bm p}^\text{int}_{d}(t) - {\bm p}\right),
	\end{equation}
	\begin{equation}
	{\dot{\bm J}}_{p,\text{int}} := \sum_{i=1}^{4}\frac{\partial \bm J_p}{\partial q_i}{\dot{q}}_{\text{int},i},
	\end{equation}
	with $k_p^\text{int} \in \mathbb{R}^+$. Please note that the velocity feedback term from \eqref{eq:e1} has dropped out in \eqref{eq:pos_task_int} since $\dot{{\bm p}}^a_{d}(\bm p) - \bm J_p \dot{\bm q}_\text{int} = \bm 0$. Similarly, the cost corresponding to the orientation task can be formulated, leading to the interim mode cost function
	\begin{equation}\label{eq:cost_int}
	\begin{aligned}
	E_\text{int} = \ & w_p\left(\ddot{\bm q}^T \bm J_{p}^T \bm J_{p} \ddot{\bm q} + 2 \left.{\bm\eta}^\text{int}_{p}\right.^T \bm J_{p} \ddot{\bm q}\right) \\ + \  & w_\theta\left( \ddot{\bm q}^T \bm J_{\theta}^T \bm J_\theta\ddot{\bm q} + 2{\eta}^\text{int}_{\theta} \bm J_\theta \ddot{\bm q}\right).
	\end{aligned}
	\end{equation}
	Regarding constraints, the equation of motion constraint \eqref{eq:EOM_ante} is modified, also replacing ${\dot{\bm q}}$ by ${\dot{\bm q}}_\text{int}$, resulting in the interim mode QP formulation 
	\begin{equation}
	(\ddot{\bm q}^*, \bm \tau^*) = \underset{\ddot{\bm q}, \bm \tau}{\operatorname{argmin}}	\ E_\text{int},
	\end{equation}
	s.t.
	\begin{align}\label{eq:EOM_interim}
	\bm M\ddot{\bm q} + \bm h(&\bm q,{\dot{\bm q}_\text{int}}) = \bm S\bm \tau,\\
	\underline{\bm \tau} \leq & \ \bm \tau \leq \bm \bar{\bm \tau}.
	\end{align}
	By using a controller that is based on the ante-impact reference, we assure a minimal jump in the desired torque at the moment of the first impact as a result of continuity in the feedforward acceleration $\bar{\ddot{\bm p}}_d^a$ and $\ddot{\theta}_d^a$. Meanwhile, the accumulation of the position feedback error results in a driving force to complete the contact until full contact is established. Stabilization in this mode provided by the dissipating contact dynamics.

	\subsection{Post-impact mode}\label{sec:post}
	
	After the final contact state is established, the control input is determined through the post-impact QP, in which it is assumed that both contacts remain closed. While the ante-impact and interim mode did not explicitly take the interaction forces $\bm \lambda$ into account, the post-impact QP does so, and hence, $\bm \lambda$ is included in the optimization variables together with $\ddot{\bm q}$ and $\bm \tau$. 
	The QP consists of a task that prescribes the end effector velocity through the reference defined in Section \ref{sec:reference_generation_post}, similar in structure to the ante-impact linear velocity task from \eqref{eq:pos_task_ante}, resulting in the error
	\begin{equation}\label{eq:pos_task_post}
	\bm e^p_p = \bm J_{p}\ddot{\bm q} + \bm\eta_p^p
	\end{equation}
	with
	\begin{equation}\label{eq:eta_post_p}
	\bm\eta^p_p = \dot{\bm J}_p\dot{\bm q} - \bar{\ddot{\bm p}}^p_{d}(\bm p) - k^p_p \left(\bar{\dot{\bm p}}^p_{d}(\bm p) - \bm J_p\dot{\bm q}\right),
	\end{equation}
	with $k^p_p \in \mathbb{R}^+$, and $\bar{\ddot{\bm p}}^p_{d}(\bm p)$ derived from $\bar{\dot{\bm p}}^p_{d}(\bm p)$ similar to \eqref{eq:ddot_p_a_d}.
	As in \cite{Steen2022}, we add a task encouraging an equal contact force distribution over both contact points, with corresponding weight $w_\lambda$, leading to the total cost function
	\begin{equation}\label{eq:cost_post}
	E_\text{post} = w_p\left(\ddot{\bm q}^T \bm J_{p}^T \bm J_{p} \ddot{\bm q} + 2\left.\bm \eta^p_p\right.^T \bm J_{p} \ddot{\bm q}\right) + w_\lambda (\lambda_1 - \lambda_2)^2.
	\end{equation}
	For the post-impact constraints, the constraint prescribing the equations of motion should now include the contact forces, hence \eqref{eq:EOM} is included in the QP. To prevent loss of contact, we enforce $\bm \lambda \geq 0$, and under the assumption that both contacts are and remain closed, we include a constraint that enforces $\ddot{\bm \gamma} = \bm 0$, resulting in the post-impact QP
	\begin{equation}
	(\ddot{\bm q}^*, \bm \tau^*, \bm \lambda^*) =  \underset{\ddot{\bm q}, \bm \tau,\bm\lambda}{\operatorname{argmin}} \	E_\text{post},
	\end{equation}
	s.t.
	\begin{align}
	\bm M \ddot{\bm q} + \bm h &= \bm S \bm \tau + \bm J_{N}^T  \bm \lambda, \\
	\bm J_N \ddot{\bm q} + \dot{\bm J}_N \dot{\bm q} &= \bm 0, \\ 
	\bm \lambda &\geq \bm 0, \\
	\underline{\bm \tau} \leq & \ \bm \tau \leq \bm \bar{\bm \tau}.
	\end{align}
	
	\section{Numerical validation}\label{sec:numerical_validation}
		
	To validate the proposed control approach, numerical simulations have been performed\footnote{All simulations can be reproduced using the publicly available Matlab scripts that can be found in \href{https://gitlab.tue.nl/robotics-lab-public/time-invariant-reference-spreading-for-simultaneous-impacts}{https://gitlab.tue.nl/robotics-lab-public/time-invariant-reference-spreading-for-simultaneous-impacts}.} tasking the robot to execute a motion containing a simultaneous impact between the plank and both contact points on the end effector. The goal is to show the robustness of the proposed control approach against uncertainties in the environment resulting in unplanned simultaneous impacts. 
	Despite the corresponding unpredictable series of velocity jumps and unknown contact states, it is shown that the proposed approach ensures execution of the desired motion without unwanted spikes in the input torque. 
	
	Simulations are performed for two different robot models. First, results are presented that use the ideal, rigid robot model with rigid contact model as described in Section \ref{sec:robot_dynamics}. After this, the simulations are repeated for a robot model with flexibility and a low-level torque control loop modeled in its joints, with contact described via a compliant contact model, as was done in \cite{Steen2022}. The latter model more closely resembles reality, as flexibility is generally present in, e.g., the drivetrain of robot joints, resulting in oscillations as a result of impacts. 
	These simulations suggest that the developed approach, which uses the assumption of a rigid robot and contact model, is suitable for real-life robot control. 
	We will use both simulation models to compare the proposed control approach against two similar baseline approaches. In the first baseline approach, referred to as the approach with \textit{no impact map}, we substitute the post-impact reference $\bar{\dot{\bm p}}^p_d(\bm p)$ by $\dot{\bm p}^p_{d,f}(\bm p)$ from \eqref{eq:p_d_f_p}, which implies that the post-impact reference is not coupled to the ante-impact reference using the impact map. In the second baseline approach, referred to as the approach with \textit{no interim mode}, we directly switch to the post-impact control mode upon detection of the first contact, instead of switching to the interim mode first. To mimic uncertainty in the environment, the configuration of the plank $q_4$ is initialized with an offset compared to the estimated configuration $q^-_{4,\text{est}}$. 

	\begin{figure}
		\centering
		\includegraphics[width=0.89\linewidth]{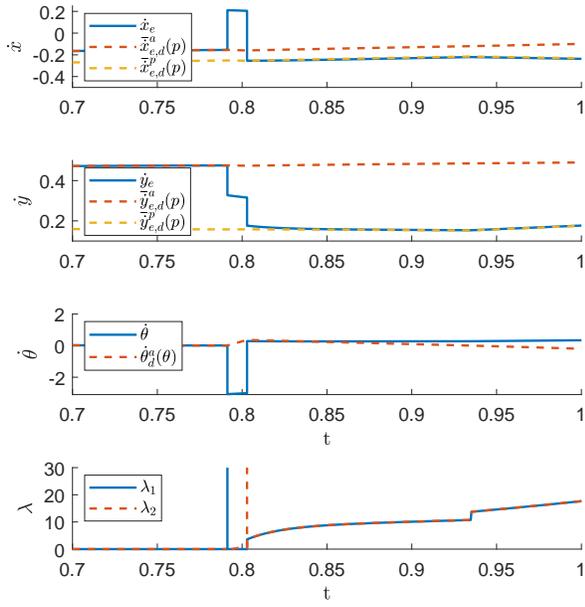}
		\caption{Cartesian end effector velocities and contact forces for the rigid robot for the proposed control approach.}
		\label{fig:velocities_rigid}
	\end{figure}
	\begin{figure}
		\centering
		\includegraphics[width=0.87\linewidth]{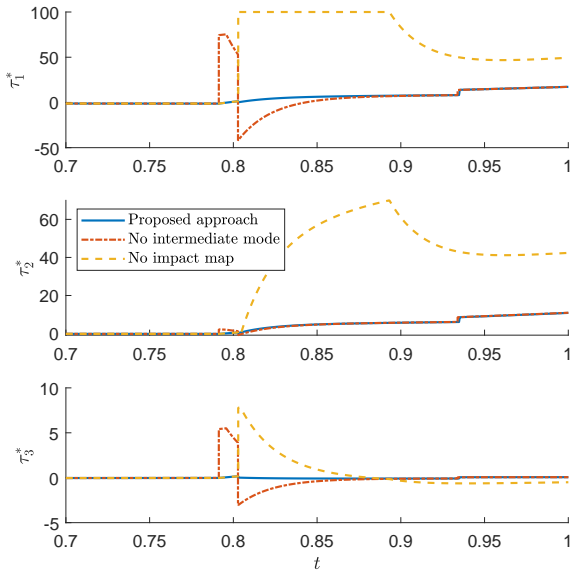}
		\caption{Commanded torque as simulated using a rigid robot model for different control approaches.}
		\label{fig:torque_rigid}
	\end{figure}
	\begin{figure}
		\centering
		\includegraphics[width=0.89\linewidth]{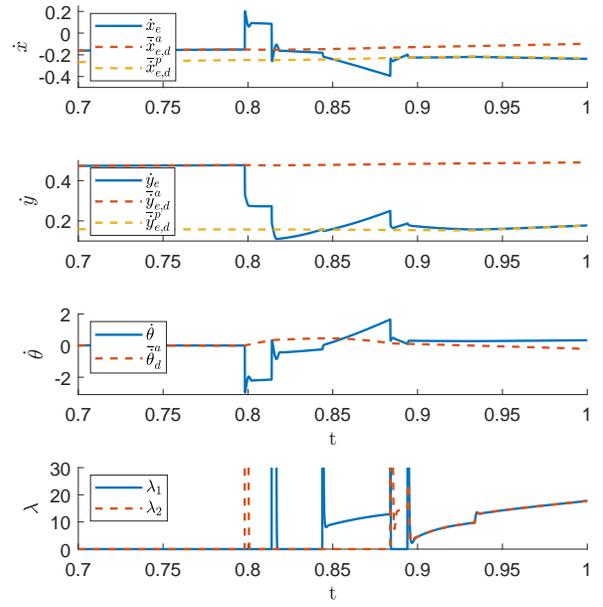}
		\caption{Cartesian end effector velocities and contact forces for the flexible robot for the proposed control approach.}
		\label{fig:velocities_flex}
	\end{figure}
	\begin{figure}
		\centering
		\includegraphics[width=0.87\linewidth]{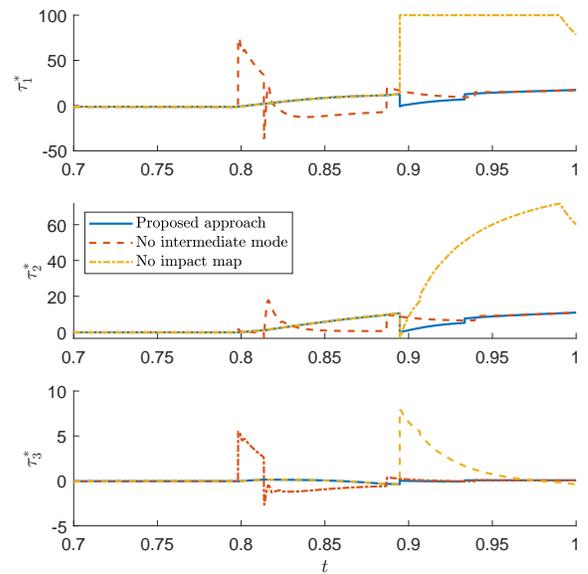}
		\caption{Commanded torque as simulated using a flexible robot model for different control approaches.}
		\label{fig:torque_flex}
	\end{figure}

	\subsection{Numerical results with a rigid robot model}\label{sec:numerical_validation_rigid}

	First, we will show the results of the simulation that uses a rigid robot model. In Figure \ref{fig:velocities_rigid}, the velocities of such simulation with the proposed control approach are presented. It can be observed that two separated impacts occur as a result of the aformentioned initial offset in $q_4$ and the corresponding misalignment between the end effector and the impact line at the moment of the first impact. Before the first impact, when the system evolves in the ante-impact mode, the reference $\bar{\dot{\bm p}}^a_d(\bm p)$ is perfectly followed as expected. After the second impact, when contact is completed and the post-impact mode is entered, it can be observed that the velocity jumps very close to the desired post-impact velocity $\bar{\dot{\bm p}}^a_d(\bm p)$, which is expected as a result of the reference formulation procedure that couples the ante-impact and post-impact reference through the impact map. Between the impacts, when the interim mode is active, the velocities, especially $\dot{x}_e$ and $\dot{\theta}$ jump far away from both the corresponding ante-impact and post-impact reference velocities (before jumping back as a result of the second impact), highlighting that no velocity reference and thus no velocity error can be defined that should be tracked in this mode. After the impacts, the linear velocity reference (note that no reference is prescribed for $\theta$ in the post-impact mode) and desired torque task described in Section \ref{sec:post} are followed as expected.

	The effect of applying velocity feedback before full contact is established can be seen in Figure \ref{fig:torque_rigid}, where the approach with no interim mode, in which the post-impact mode is entered upon first contact detection, leads to a sudden increase in input torque $\bm \tau^*$ after the first impact occurs. This is caused by a large error between the actual velocity and the post-impact reference velocity. As soon as full contact is established, the torque jumps in opposite direction, to reduce the velocity error caused by the unwanted effort of the controller before full contact is established. This is clearly undesirable behavior, which the proposed approach avoids by the novel interim mode design. 
	
	When the post-impact vector field is not constructed by means of the impact map, switching to the post-impact mode after the final impact is detected can lead to a large velocity tracking error, which can subsequently lead to a large input torque $\bm \tau^*$ as can be observed from Figure \ref{fig:torque_rigid}, where $\tau_1^*$ even reaches the maximum joint torque set in $\bar{\bm \tau}$. This can have an undesired effect, as the sudden extreme jump in $\bm \tau^*$ torque can trigger unwanted effects in terms of performance and robustness. 
	As opposed to the two baseline approaches, almost no jump in $\bm \tau^*$ can be observed upon occurrence of either of the two impacts in the proposed control approach, which highlights the benefits of this approach.

	\subsection{Numerical results with a flexible robot model}
	
	To validate the effectiveness of the control strategy in a more realistic setting, simulations are also performed on a model of the robotic manipulator from Figure \ref{fig:robot_3DOF} that more closely matches reality, identical to the flexible robot model used for validation in \cite{Steen2022}. In this model, the transmission in each joint is modeled as an element with finite stiffness, damping and inertia, and an additional low-level torque control loop is implemented as described in \cite{Albu2007}, while interaction between the robot and the impacting surface is computed using the compliant exponentially extended Hunt-Crossley contact model from \cite{Carvalho2019}. 	

	In Figure \ref{fig:velocities_flex}, the results of simulations with this model are depicted, showing clearly that contact is broken and reinstated several times as a result of the compliant partially elastic contact model, while the flexibility in the joints causes internal vibrations shortly after each impact.  
	However, despite the presence of these vibrations after the initial impact, the desired torque $\bm \tau^*$ resulting from the interim mode QP in the proposed approach, seen in Figure \ref{fig:torque_flex}, does not show any of these vibrations, while sustained contact is still established in the end. This is an additional advantage on top of the benefits described in Section \ref{sec:numerical_validation_rigid}, resulting from the fact that the velocity measurements $\dot{\bm q}$ are not used in the interim mode. This benefit is highlighted when comparing the proposed approach with the baseline approach with no interim mode, where, aside from the jump in $\bm \tau^*$ already observed in simulations with the rigid model, the internal vibrations cause rapid fluctuations in $\bm \tau^*$, which can induce additional undesirable vibrations into the system. 
	
	After the interim mode is finished and full contact is established, a small jump in $\bm \tau^*$ can be observed for the proposed approach, which is natural given the additional flexibility that is not explicitly described in the QP controllers. However, when comparing the proposed approach with the approach with no impact map, we can see that the corresponding jump in $\bm \tau^*$ is much larger, increasing the likelihood of unwanted effects in terms of performance and robustness.

	\section{Conclusion}\label{sec:conclusion}

	In this paper, a new control framework based on QP control is proposed for robotic manipulators that experience nominally simultaneous impacts. This framework uses separate time-invariant reference fields to prescribe the ante-impact and post-impact tasks, which are coupled through an inelastic impact map to minimize the velocity error upon switching. A distinctive feature of the control framework is the additional interim mode, during which position feedback is applied based on time-integration of the ante-impact reference velocity field, in order to establish full sustained contact.	
	Simulation results on a model of a rigid robot with inelastic contact model, and a more realistic flexible robot model with compliant partially elastic contact model, show the effectiveness of the approach. For both models, the motion is successfully executed with accurate tracking of the desired velocity, while minimizing unwanted spikes or fluctuations in the input torque.  
	Future work involves scaling up to a more realistic 3D case and performing a corresponding study on real-life robotic system. This also includes a further study in the moment of switching between control modes based on impact detection. It does, however, not require methodological alterations to the presented approach.

\addtolength{\textheight}{-0cm}   



\section*{APPENDIX}


	\subsection{Formulation ante-impact reference} \label{app:appendix_reference_ante}

	The ante-impact linear velocity reference ${\dot{\bm p}}^a_d(\bm p)$ is formulated by means of a constant user-defined impact position and velocity $\bm p_\text{imp}$ and $\bm v_\text{imp}$, which indicate the center position of the targeted set of positions on the nominal impact line and the corresponding desired ante-impact velocity. A target position $\bm p_t(\bm p)$ is defined, which is the position at equal distance from $\bm p_\text{imp}$ as the distance between $\bm p$ and $\bm p_\text{imp}$, but aligned to hit $\bm p_\text{imp}$ with velocity $\bm v_\text{imp}$, through
	\begin{equation}\label{eq:p_t}
		\bm p_t(\bm p) = \bm p_\text{imp} -  \frac{\bm v_\text{imp}\norm{\bm p_\text{imp} - \bm p}_2}{\norm{\bm v_\text{imp}}_2}.
	\end{equation}
	The desired ante-impact velocity ${\dot{\bm p}}^a_d(\bm p)$ is then formulated as a combination of $\bm v_\text{imp}$ and a position feedback term using $\bm p_t(\bm p)$, scaled such that for all $\bm p$, the norm of ${\dot{\bm p}}^a_d(\bm p)$ is equal to that of $\bm v_\text{imp}$. This gives
	\begin{equation}
		{\dot{\bm p}}^a_d(\bm p) = \frac{\bm v_\text{imp} + \alpha(\bm p_t(\bm p) - \bm p)}{\norm{\bm v_\text{imp} + \alpha(\bm p_t(\bm p) - \bm p)}_2}\norm{\bm v_\text{imp}}_2,
	\end{equation}
	with $\alpha \in \mathbb{R}^+$ as a user-defined parameter, where increasing $\alpha$ will result in a reference velocity field that steers impacts closer to $\bm p_\text{imp}$ by taking a wider path towards the nominal impact line.

	\subsection{Formulation post-impact reference} \label{app:appendix_reference_post}

	To formulate the post-impact reference contribution $\bar{\dot{\bm p}}^p_{d,\text{imp}}(\bm p)$ corresponding to the nominal post-impact velocity, we first sample $N$ points that are evenly distributed on the nominal impact line, captured by $\bm p_{\text{nom},i}$ with $i \in \{1, \dots, N\}$. Assuming that the end effector is perfectly aligned with the impact line ($\theta = q_4$) in a non-singular configuration, we can then use inverse kinematics to create a set $\bm q_{\text{nom},i}$ with the joint configurations corresponding to $\bm p_{\text{nom},i}$. Using inverse velocity kinematics and assuming to have an estimation of the plank position $q^-_{4,\text{est}}$ and velocity $\dot{q}^-_{4,\text{est}}$ at the moment of impact, we can determine the corresponding nominal ante-impact joint velocities $\dot{\bm q}^-_{\text{nom},i}$, as
	\begin{equation}
	\dot{\bm q}^-_{\text{nom},i} = \left[\left.\dot{\bm q}_{\text{rob},\text{nom},i}^-\right.^T \ \dot{q}^-_{4,\text{est}}\right]^T,
	\end{equation}
	with
	\begin{equation}
	\dot{\bm q}_{\text{rob},\text{nom},i}^- = \begin{bmatrix}\bm J_{p,\text{rob}}(\bm q_{\text{nom},i}) \\ \bm J_{\theta,\text{rob}}(\bm q_{\text{nom},i})\end{bmatrix}^{-1}  \begin{bmatrix} \bar{\dot{\bm p}}^a_{d}(\bm p_{\text{nom},i}) \\ \dot{{\theta}}^a_{d}(q^-_{4,\text{est}})\end{bmatrix}.
	\end{equation}
	Using the impact map \eqref{eq:impact_map}, we can determine the corresponding post-impact joint velocities $\dot{\bm q}_{\text{nom},i}^+$, which can be translated to velocities in operational space via
	\begin{equation}
	{\dot{\bm p}}^p_{d,\text{imp}}(\bm p_{\text{nom},i}) = \bm J_p(\bm q_{\text{nom},i})\dot{\bm q}_{\text{nom},i}^+.
	\end{equation}
	To ensure that we can compute ${\dot{\bm p}}^p_{d,\text{imp}}(\bm p)$ for each position without having to recompute the inverse kinematics and impact map, we interpolate between the values of $\dot{\bm p}^p_{d,\text{imp}}(\bm p_{\text{nom},i})$ using radial basis function interpolation. Finally, we extend this velocity field in the directions normal to the nominal impact line in similar fashion to the extension in \eqref{eq:extension_ante}, to come to the expression of $\bar{\dot{\bm p}}^p_{d,\text{imp}}(\bm p)$ as
	\begin{equation}
		\bar{\dot{\bm p}}^p_{d,\text{imp}}(\bm p) := {\dot{\bm p}}^p_{d,\text{imp}}(\bm p + d(\bm p) \bm n_\text{imp}),
	\end{equation}
	recalling that $d(\bm p)$ represents the smallest distance from $\bm p$ to the nominal impact line, while $\bm n_\text{imp}$ represents the vector normal to the nominal impact line, as seen in Figure \ref{fig:reference}.

	\section*{ACKNOWLEDGMENT}

	This work was partially supported by the Research Project I.AM. through the European Union H2020 program under GA 871899.

\bibliography{References/library}{}
	\bibliographystyle{ieeetr}


\end{document}